\documentclass[a4paper]{article}

\pdfoutput=1
\usepackage[utf8]{inputenc}

\usepackage[super,sort&compress,comma]{natbib}

\usepackage[english]{babel}
\usepackage[T1]{fontenc}
\usepackage{csquotes}

\usepackage[a4paper,top=3cm,bottom=2cm,left=3cm,right=3cm,marginparwidth=1.75cm]{geometry}

\usepackage{amsmath}
\usepackage{graphicx}
\usepackage{hyperref}
\hypersetup{colorlinks=true, allcolors=blue}
\usepackage{authblk}
\usepackage{multirow}

\usepackage{listings}
\usepackage{xcolor}

\definecolor{codegreen}{rgb}{0,0.6,0}
\definecolor{codegray}{rgb}{0.5,0.5,0.5}
\definecolor{codepurple}{rgb}{0.58,0,0.82}
\definecolor{backcolour}{rgb}{0.95,0.95,0.92}

\lstdefinestyle{mystyle}{
    backgroundcolor=\color{backcolour},   
    commentstyle=\color{codegreen},
    keywordstyle=\color{magenta},
    stringstyle=\color{codepurple},
    basicstyle=\ttfamily\scriptsize,
    breakatwhitespace=false,         
    breaklines=true,                 
    captionpos=b,                    
    keepspaces=true,                 
    showspaces=false,                
    showstringspaces=false,
    showtabs=false,                  
    tabsize=2
}

\usepackage{nomencl}
\usepackage{booktabs}
\usepackage{tabularx}
\usepackage{textcomp} 
\usepackage{amsfonts}
\usepackage{outlines} 
\usepackage{subfig}
\usepackage[version=4]{mhchem}
\usepackage{siunitx}
\usepackage{bm}
\usepackage[normalem]{ulem} 
\date{}

\DeclareSIUnit\angstrom{\text {Å}}

\usepackage[final]{changes}

\title{PAL - Parallel active learning for machine-learned potentials}

\author[1,2]{Chen Zhou}
\author[1,2]{Marlen Neubert}
\author[1,2]{Yuri Koide}
\author[1,2]{Yumeng Zhang}
\author[1,2,3]{Van-Quan Vuong}
\author[1,2]{Tobias Schlöder}
\author[2]{Stefanie Dehnen}
\author[1,2,*]{Pascal Friederich}

\affil[1]{Institute of Theoretical Informatics, Karlsruhe Institute of Technology, Kaiserstr. 12, 76131 Karlsruhe, Germany}
\affil[2]{Institute of Nanotechnology, Karlsruhe Institute of Technology, Kaiserstr. 12, 76131 Karlsruhe, Germany}
\affil[3]{Institute for Physical Chemistry, Karlsruhe Institute of Technology, Kaiserstr. 12, 76131 Karlsruhe, Germany}
\affil[*]{Corresponding author: pascal.friederich@kit.edu}

\begin{document}

\maketitle
\begin{abstract}
    Constructing datasets representative of the target domain is essential for training effective machine learning models. Active learning (AL) is a promising method that iteratively extends training data to enhance model performance while minimizing data acquisition costs. However, current AL workflows often require human intervention and lack parallelism, leading to inefficiencies and underutilization of modern computational resources. In this work, we introduce PAL, an automated, modular, and \textbf{p}arallel \textbf{a}ctive \textbf{l}earning library that integrates AL tasks and manages their execution and communication on shared- and distributed-memory systems using the Message Passing Interface (MPI). PAL provides users with the flexibility to design and customize all components of their active learning scenarios, including machine learning models with uncertainty estimation, oracles for ground truth labeling, and strategies for exploring the target space. We demonstrate that PAL significantly reduces computational overhead and improves scalability, achieving substantial speed-ups through asynchronous parallelization on CPU and GPU hardware. Applications of PAL to several real-world scenarios - including ground-state reactions in biomolecular systems, excited-state dynamics of molecules, simulations of inorganic clusters, and thermo-fluid dynamics - illustrate its effectiveness in accelerating the development of machine learning models. Our results show that PAL enables efficient utilization of high-performance computing resources in active learning workflows, fostering advancements in scientific research and engineering applications.

\end{abstract}

\section{Introduction}\label{sec:introduction}

In many ML applications, the primary goal of data generation is to create a diverse dataset that comprehensively represents the relevant data distribution, ensuring stability and reliability during deployment\cite{bejani2021systematic}. In contrast, scientific applications often require ML models to operate in exploratory settings, where discovering novel and unseen inputs is expected or even desired. ML models trained on limited initial datasets often struggle to generalize effectively to newly encountered instances outside the training distribution. Continuously updating the model by labeling portions of the explored input space and retraining is inefficient, as it frequently adds redundant data points to the training set without significantly improving generalization accuracy. To address this challenge, active learning (AL) has been introduced as an efficient approach to building accurate and reliable models while minimizing queries to an oracle -- a computational method or experimental process capable of providing ground truth labels \cite{jordan2015data}. By iteratively selecting the most informative instances for labeling, AL reduces the overall cost of training set generation and enhances model performance \cite{settles2009active,settles2011theories,ren2021survey}.

\begin{figure}
  \centering
  \includegraphics[width=\textwidth]
  {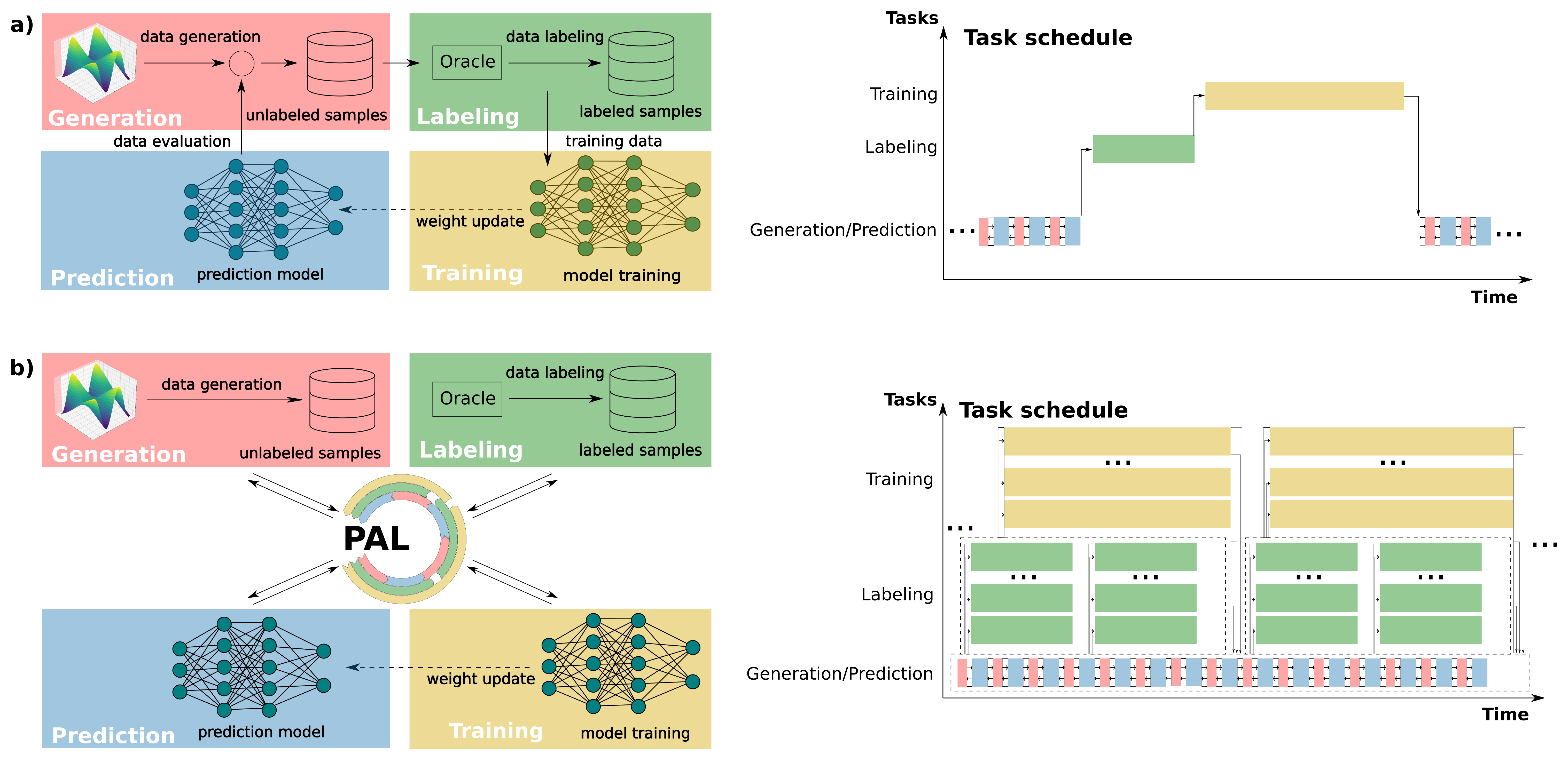}  
  \caption{Comparison of (a) a conventional (serial) active learning and (b) our parallel active learning workflow PAL. a) Classical active learning workflow, in which different tasks , i.e. exploration of the input space using generation and prediction kernels, labeling of the samples using the oracle kernel, and training of the ML model, are performed iteratively and sequentially. b) PAL modularizes, decouples, and parallelizes data generation, oracle labeling, and ML training processes.}
  \label{fig:title}
\end{figure}

One particularly important application of active learning is developing new machine-learned potentials, which are often trained on quantum mechanical calculations to map three-dimensional atomic coordinates to corresponding total energies and forces \cite{behler2007generalized,bussi2020using,friederich2021machine,behler2021four,deringer2021gaussian}. These potentials are ML models typically based on neural networks \cite{batzner20223,batatia2023foundation,chen2023discovering,donkor2023mlmdwater,topel2023learned,sasmal2023reaction,akher2023semiclassical,hui2023cyclic,vandenhaute2023machine} or Gaussian process models \cite{deringer2021gaussian,vandermause2020algp,vandermause2022active,duschatko2024uncertainty}. The advantage of ML potentials lies in their ability to perform molecular dynamics (MD) simulations with the accuracy of ab initio calculations but at only a fraction of the computational cost. As MD simulations often aim to explore the behavior and function of molecules and materials at the atomistic scale, constructing an initial dataset that fully represents all relevant geometries encountered during simulations is nearly impossible. Active learning approaches are therefore essential to update datasets and machine-learned potentials iteratively, ensuring both reliability and accuracy throughout the simulations \cite{smith2018less,jinnouchi2020fly,vandermause2020fly,wang2020active,li2021automatic,saleh2021active,ang2021active,smith2021automated,young2022reaction,vandermause2022active,farache2022active,kleiman2023active,li2024active,zhang2024modelling}. Typically, active learning for machine-learned potentials is performed in a batch-wise manner, iterating through cycles of data labeling via quantum chemistry calculations, training ML models, and ML-accelerated simulations or other sampling strategies. Uncertainty quantification \cite{kahle2022quality,kulichenko2023uncertainty,duschatko2024uncertainty} is often used to select the most informative samples for subsequent iterations (see Figure~\ref{fig:title}a).

Despite the widespread use of active learning for machine-learned potential development, the current infrastructures and algorithms often fail to fully utilize modern computational resources, relying heavily on human intervention for tasks such as labeled data handling and model training initialization. Furthermore, the lack of parallelism introduces inefficiencies, as active learning is typically performed sequentially; for example, labeled data generation, model training, and exploration in molecular dynamics (MD) simulations are executed one after another.

To address these challenges, we developed PAL, an automated, modular, and parallel active learning workflow (see Figure~\ref{fig:title}b) with several key advantages: (1) the fully automatic workflow minimizes human intervention during execution; (2) the modular and highly adaptive design reduces the effort of (re-)implementing parts of the active learning workflow while allowing it to be extended to various tasks with diverse combinations of resources, data, and ML model types; (3) the decoupling of all AL modules enables data and task parallelism, facilitating simultaneous exploration/generation, labeling, and training tasks; and (4) PAL is implemented using the Message Passing Interface (MPI) in Python \cite{mpi, mpi4py2021, mpipy2023}, ensuring scalability and flexibility for deployment on both shared-memory systems (e.g., laptops, local workstations) and distributed-memory systems (e.g., high-performance computing clusters).

We demonstrate the applicability and versatility of PAL across various applications of machine-learned potentials and other machine learning tasks for science and engineering beyond atomistic simulations. Our results highlight the scalability of PAL and the significant speed-ups achieved through asynchronous parallelization on CPU and GPU hardware.

\section{Description of the PAL workflow}
\begin{figure}
  \centering
  \includegraphics[width=\textwidth]{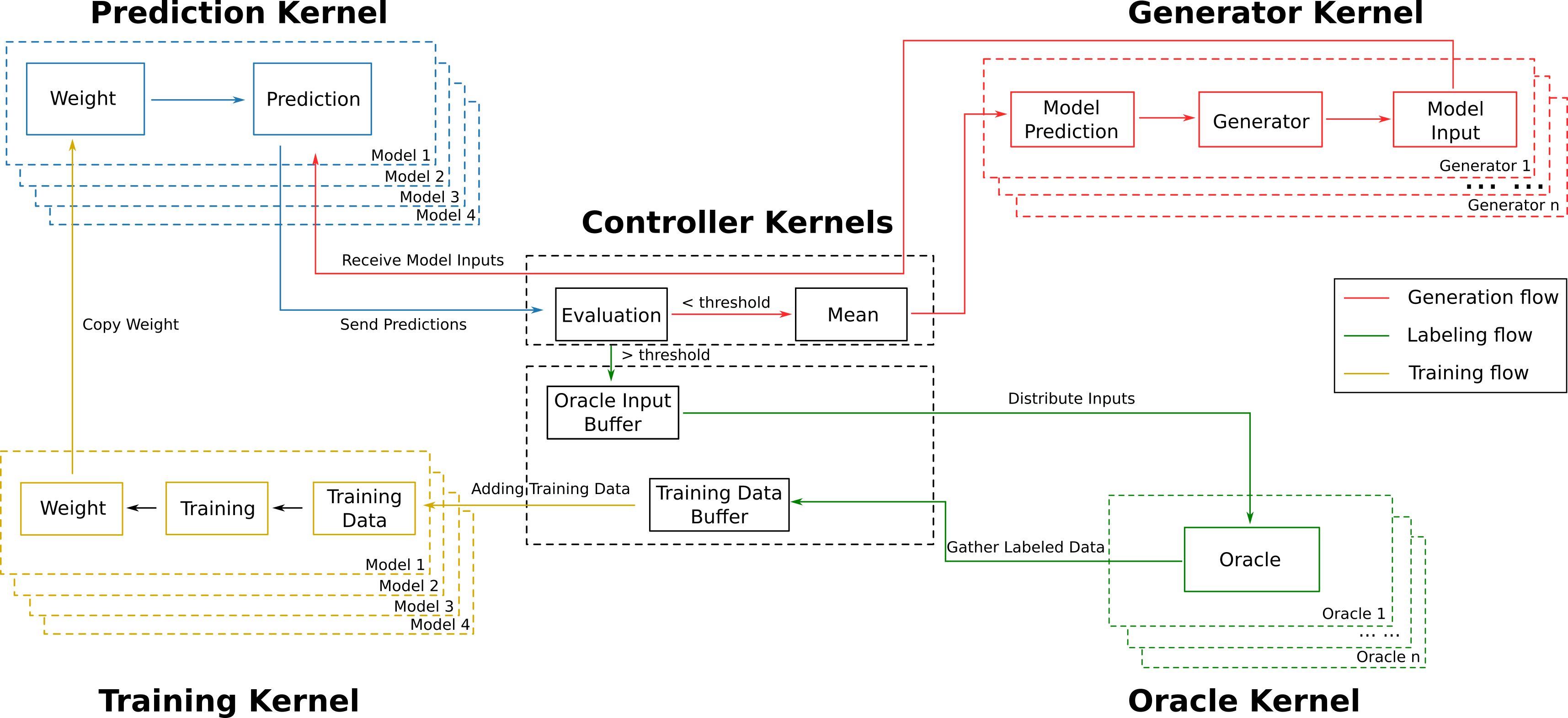}
  \caption{The computational architecture of the PAL workflow. Multiple boxes indicate parallelization of multiple instances of each kernel. The arrows illustrate information flow between the kernels orchestrated by the two controller sub-kernels. One dedicated controller sub-kernel ensures high-frequency communication between generation and prediction kernels.
  }
  \label{fig:al}
\end{figure}

PAL is designed as a modular and scalable framework that orchestrates the components of an active learning workflow in a parallel computing environment. The architecture of PAL centers around five core modules, referred to as kernels: the prediction kernel, generator kernel, training kernel, oracle kernel, and controller kernel (see Figure~\ref{fig:al}). These kernels operate concurrently and communicate efficiently through the Message Passing Interface (MPI), enabling seamless integration and coordination on both shared- and distributed-memory systems.

At a high level, the prediction kernel provides machine learning models that make predictions on input data generated by the generators. The generator kernel explores the target space by producing new data instances based on the predictions received. The training kernel retrains the machine learning models using newly labeled data to improve their performance iteratively. The oracle kernel acts as the source of ground truth labels, providing accurate labels for selected data instances that require further clarification. Overseeing these components, the controller kernel manages the workflow by coordinating communication, scheduling tasks, and handling data exchanges between kernels.

This modular design allows users to customize and extend each kernel according to the specific requirements of their active learning scenarios. By decoupling the kernels and enabling them to operate in parallel, PAL maximizes computational resource utilization and minimizes overhead associated with sequential processing. The architecture supports asynchronous execution, where data generation, model prediction, labeling, and training can proceed simultaneously, leading to significant improvements in efficiency and scalability. A user-designated number of instances of the process in every kernel are generated during execution, each with distinct identifiers, data, and behaviors, tailored to the specific requirements of the kernel. The functionality of each kernel will be discussed in more detail in the subsequent sections.

\subsection{Prediction kernel}

In the prediction kernel, ML processes infer target values for inputs generated by the generators. Multiple ML models can operate concurrently when bootstrapping or query-by-committee techniques are employed. The controller aggregates their predictions and performs predefined or user-defined manipulations, such as calculating the mean and standard deviation, before distributing the results to generators and oracles. The separation of prediction and training tasks in PAL aims to minimize disruptions in data generation and inference caused by the time-consuming labeling and training processes. ML models in the prediction kernel are updated periodically by replicating weights from the corresponding models in the training kernel after a specified number of training epochs.

\textit{Example:} In the case of machine-learned potentials, the prediction kernel is the machine learning model that predicts energies and forces given the system coordinates during an MD simulation or during other sampling methods used to explore the relevant geometry space (e.g. enhanced sampling methods or transition state search algorithms). Potential prediction kernels can be SchNet\cite{schutt_schnet_2018}, Nequip\cite{batzner20223}, Allegro\cite{musaelian_learning_2023}, MACE\cite{Batatia2022mace}, and other machine-learned potentials.

\subsection{Generator kernel}

The generator kernel hosts an arbitrary number of generator processes running in parallel to accelerate data generation. Predictions from the ML models in the prediction kernel are disseminated to each generator by the controller, facilitating further data production. Each generator independently manages its data and maintains its generation-prediction-iteration status. Every generator can signal the controller kernel to shut down the PAL workflow upon meeting user-defined criteria.

\textit{Example:} In the case of machine-learned potentials, this kernel is the exploration algorithm, e.g. a single MD step or the generation of a new geometry in the geometry-exploration method. The generator kernel communicates with the prediction kernel through the controller kernel to receive energy and force labels for given geometries, e.g. to propagate the system and propose new geometries. Neither the generator kernel nor the prediction kernel makes any decisions about the ML predictions' reliability, as the controller handles this process centrally. However, the generator kernel receives reliability information from the controller in order to decide whether to trust the energies and forces predicted by the ML models or whether to restart trajectories. That means that the uncertainty quantification and thus the decision of whether or not to label new geometries by the oracle is handled centrally by the controller kernel whereas the decision-making logic of how to react to uncertain predictions is implemented by the generator kernel. This offers flexibility and allows users to implement a wide range of workflows, e.g. allowing trajectories to propagate into regions of high uncertainty for a given number of steps ('patience').

\subsection{Oracle kernel}

The oracle kernel allows for deploying multiple oracle processes, each operating independently. Every oracle process keeps point-to-point communication with the controller, receiving input data for labeling. Once labeling is complete, the generated labels sets are returned to the controller before being distributed to the training kernel.

\textit{Example:} In the case of machine-learned potentials, this kernel is the quantum chemical calculation that is used to generate the labels for the training data, e.g. density functional theory calculations to compute energies and forces of given input geometries. As described above, the decision about when to invoke additional DFT calculations to label data is performed centrally by the controller kernel. Once the active learning workflow is 'converged', i.e. the entire relevant input space is covered by the dataset and the uncertainty of the trained ML model on the final dataset does not reach a certain threshold value anymore, no new oracle calls will be requested anymore and PAL will run simulations by only iterating between generator and prediction kernels.

\subsection{Training kernel}

The training sets kept in the training kernel are expanded with new data labeled by oracles. An equal number of ML models as in the prediction kernel are trained in parallel within the training kernel, synchronizing with the labeling and prediction-generation processes. Training can be halted if a user-defined early stopping criterion is met to prevent overfitting, or restarted when new data points are introduced. For efficiency, trained model weights are periodically copied directly to the prediction kernel. The training kernel manages all model-related data, including scalars, hyperparameters, weights, and training histories, as models in the prediction kernel are considered replicas of those in the training kernel. Every process in the training kernel can signal the controller kernel to shut down the PAL workflow upon meeting user-defined criteria.

\textit{Example:} In the case of machine-learned potentials, as discussed in the paragraph on prediction kernels above, the training kernel includes one epoch of training of any machine-learned potential, e.g. SchNet, Nequip, Allegro, or MACE, on a given data set. The user can define whether training should continue from the previous checkpoint or restart from a new random initialization after a certain number of epochs and active learning iterations\cite{ash2020warmstartingneuralnetworktraining}. PAL offers full flexibility for the user. Generally, we recommend not to re-initialize weights during active learning which is also the default setting. 


\subsection{Controller Kernel}

The controller kernel orchestrates data communication, evaluates model predictions, selects inputs for labeling, and manages metadata storage (oracle input buffer and training data buffer). Data selected for labeling is buffered in the oracle input buffer and sent to the first available oracle. Labeled data is stored in the training data buffer and distributed to the ML models in the training kernel once the buffer size reaches a user-defined threshold. The data flow between the prediction and generator kernels is decoupled from the oracle and training kernels, ensuring efficient and uninterrupted communication between the prediction and generator kernels. The oracle and training kernels can be disabled to convert PAL into a prediction-generation workflow without significant performance impact, useful in scenarios where model training is unnecessary, such as ML-accelerated molecular dynamic simulations.

The user does not have to add any code to the controller kernel, except to provide functions that select instances for labeling and adjust the training data buffer dynamically (see \textbf{Utilities} in the SI for more detail). Other than that, the user only needs to specify and adjust the previous four kernels for prediction, generation, oracle, and training.

\section{Illustrating Applications}
The PAL library developed and presented in this work has been applied in several scenarios in different application areas, also beyond atomistic simulations. The examples discussed in the following include photodynamics simulations using surface hopping algorithms based on neural network potentials and TD-DFT\cite{zhou2023active}, hydrogen transfer reaction simulations in biological systems using graph neural network potentials trained on semiempirical methods as well as DFT\cite{riedmiller2024substituting}, simulations of inorganic clusters using neural networks trained on DFT data, as well as surrogate machine learning models of fluid- and heat transport in textured channels trained on fluid dynamics simulation data\cite{kaithakkal2023heat,koide2024machine}. By altering the kernels (see Table~\ref{tab:applications}), the library is adaptive to the different resource requirements, data structures, ground-truth oracles, and machine learning model types and architectures required by various scenarios.

Due to the modularity of the PAL software architecture, generic code for each of the kernels can be customized by the user to accommodate the needs of the respective application areas. All generic technical aspects of communication are identical in all applications and thus transferrable. Clear interface definitions make it easy for users to implement custom kernels by integrating their own code in the user-defined kernel functions or by calling external software from the kernels. Users can contribute kernels or kernel blueprints to the PAL code, which will make it easier for future users to use those kernels and further customize them. Examples include different quantum chemistry codes as oracle kernels, different ML potentials for training and prediction kernels, and different molecular dynamics or enhanced sampling propagators as generators.

\begin{figure}
  \centering
  \includegraphics[width=\textwidth]
  {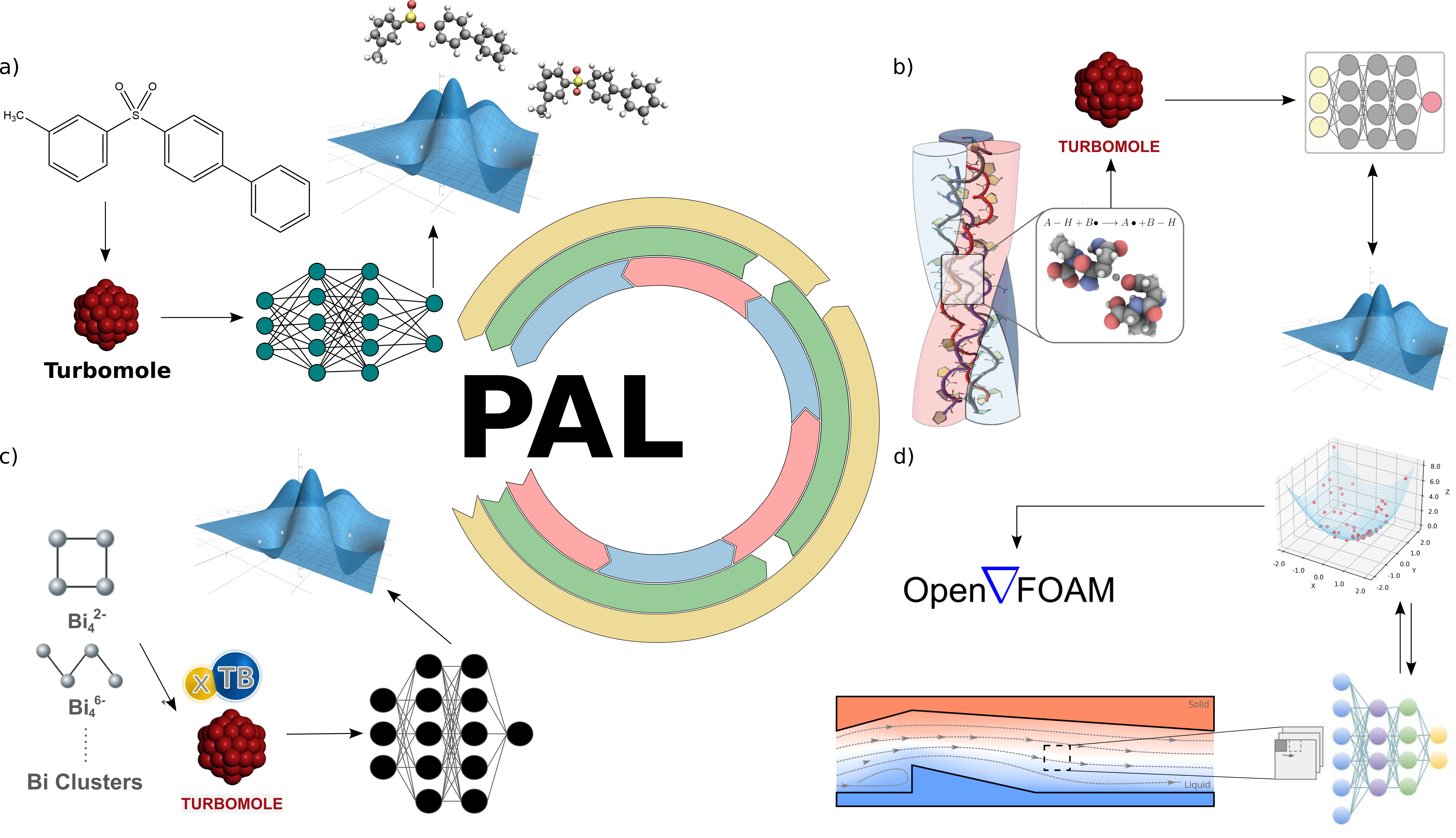}  
  \caption{Examples of PAL applications. a) Photodynamics simulations. b) Hydrogen atom transfer reaction simulations. c) Atomistic simulation of inorganic clusters. d) Thermo-fluid flow properties optimization.}
  \label{fig:app}
\end{figure}
\subsection{Photodynamics simulations}

Organic semiconductors are essential materials in both emerging and commercially significant applications, such as organic solar cells and organic light-emitting diodes (OLEDs)\cite{pollice2021organic}. While computational methods for predicting molecular properties using physics-based models\cite{friederich2017rational} as well as machine learning models\cite{friederich2019toward,reiser2021analyzing,kunkel2021active,tan2023machine,johnson2024discovery} are well developed, simulating complex phenomena like degradation remains challenging. This difficulty arises due to insufficient data to train machine learning (ML) models for these processes and the high computational cost of long excited-state dynamics simulations required by physical models.

To address this challenge, we employ PAL to enable the simulation of multiple excited-state potential energy surfaces of a small molecule organic semiconductor, 3-Methyl-4'-phenyl-diphenylsulfone. In this application of PAL, fully connected neural networks (NNs) are utilized in both the \textbf{prediction and training kernels} (see Fig.~\ref{fig:app}a). The prediction kernel leverages these NNs to approximate excited-state energies and forces efficiently.

For uncertainty quantification, we implement the query-by-committee method\cite{Seung1992qbc}, where four NN models in the prediction kernel perform energy and force predictions for the same set of molecular geometry inputs generated by the \textbf{generator kernel}. The generator kernel runs 89 molecular dynamics simulations in parallel, each exploring different regions of the chemical space to discover unseen molecular geometries. The mean predictions from the committee are used to propagate the molecular dynamics trajectories.

When the standard deviation among the committee's predictions for a given geometry exceeds a predefined threshold, indicating high uncertainty, the corresponding geometries are forwarded to the oracle kernel. In the \textbf{oracle kernel}, accurate energy and force labels are computed using time-dependent density functional theory (TDDFT) at the B3LYP/6-31G* level of theory. These new data points are then added to the training set in the training kernel, where the NNs are retrained to improve their predictive accuracy.

We deployed this workflow on two hybrid CPU-GPU nodes of the HoreKa cluster. The forward pass of 89 geometries in parallel takes an average of 51.5 ms for each NN in the prediction kernel, while MPI communication and trajectory propagation require only 4.27 ms. Notably, removing the oracle and training kernels does not affect this result, indicating that the additional communication and data processing do not degrade the performance of the rate-limiting step.

By leveraging PAL, we achieve substantial computational savings compared to traditional sequential workflows. In this application, the TDDFT calculations are the computational bottleneck, which is reduced by PAL through parallelizing TDDFT across many CPU cores as well as running model training and MD simulations in parallel. Furthermore, avoiding similar and thus redundant TDDFT calculations, we benefit from the ability to run multiple molecular dynamics simulations in parallel, exploring different parts of the input space simultaneously and thus suggesting more diverse samples to be labeled by TDDFT. This accelerates the input space exploration and thus the development of reliable ML models for photodynamics simulations, facilitating the study of complex phenomena such as degradation in organic semiconductors.
\subsection{Hydrogen atom transfer reaction simulations}
Mechanical stress in collagen can lead to bond ruptures within the protein backbone, resulting in the formation of two radicals. These radicals, characterized by unpaired valence electrons, are highly reactive and can potentially cause damage to the surrounding biological environment. Hydrogen atom transfer (HAT) is a fundamental process in radical chemistry, wherein a hydrogen atom is abstracted from a donor molecule to an acceptor, generating a new radical species. Recent research on collagen has identified HAT processes as a critical mechanism for radical migration, playing a key role in mitigating damage caused by radicals produced through mechanical stress on collagen fibrils \cite{Zapp.2020}. Due to their short lifetimes, these radicals are challenging to observe experimentally, necessitating computational approaches for their study.

Traditional classical molecular dynamics (MD) simulations and even reactive force fields struggle to accurately describe radicals and HAT processes in proteins, primarily because they cannot adequately capture the quantum mechanical nature of bond breaking and formation. To overcome this limitation, we rely on machine-learned potentials to compute molecular energies and predict reaction barriers with high accuracy. This approach enables large-scale hybrid MD and kinetic Monte Carlo simulations, providing deeper insights into the mechanisms of radical migration and HAT reactions in biological systems \cite{rennekamp_hybrid_2020}.

Constructing an effective training dataset for the machine-learned potentials presents several challenges. The potential energy surface (PES) of HAT reactions is complex, requiring sampling not only near-equilibrium configurations but also a diverse set of reaction pathways and transition states. Moreover, the trained model needs to generalize to unseen systems, such as different radical positions in unseen collagen environments, which necessitates training on various peptide combinations and reaction sites. Additionally, quantum chemical calculations to obtain ground truth energies and forces are computationally expensive, making efficient data generation and selection of informative training data essential.

In this application of PAL, the \textbf{prediction kernel} consists of a graph neural network (GNN). We utilize models such as SchNet \cite{schutt_schnet_2018}, Allegro \cite{musaelian_learning_2023}, and MACE \cite{Batatia2022mace}, which have demonstrated high accuracy in predicting energies and forces for molecular systems. The GNN in the prediction kernel provides rapid inference of energies and forces for new configurations generated during simulations (see Fig.~\ref{fig:app}b).

The \textbf{generator kernel} employs a workflow that continuously generates new reaction data points. This is achieved through methods such as molecular dynamics simulations biased towards reaction coordinates or employing transition state search algorithms to explore possible reaction pathways. By generating a stream of diverse configurations, the generator kernel effectively samples the relevant configuration space for HAT reactions.

For labeling the most informative and uncertain configurations, the \textbf{oracle kernel} uses the quantum chemistry software Turbomole \cite{ahlrichs_electronic_1989} to perform calculations at the DFT/BMK/def2-TZVPD level of theory. These calculations provide accurate ground truth energies and forces, which are essential for refining the machine-learned potentials.

In the training kernel, we leverage pre-trained models as a starting point, incorporating information from previously generated initial datasets. The \textbf{training kernel} updates these models using the new labeled data through query-by-committee uncertainty quantification, to generalize better to unseen systems and reaction sites.

In this application example, we can generate an infinite stream of diverse unlabeled samples, moving the bottleneck to the labeling process by the oracle and the training process. Depending on the specific application, rather inexpensive oracles, e.g. xTB \cite{xtb} might be sufficient, shifting the bottleneck to the training. In such cases, by parallelizing all components within the PAL framework, we efficiently cover the relevant configuration space and benefit from a continuously updated ML model which is essentially trained on an infinite stream of newly generated data \cite{schopmans2023neural}, preventing overfitting and ensuring optimal generalization. Here, this achieves accurate predictions of HAT reaction barriers with chemical accuracy. This approach accelerates the development of reliable machine-learned potentials for simulating complex biochemical reactions, ultimately enhancing our understanding of radical-mediated processes in biological systems.

\subsection{Atomistic simulation of inorganic clusters}
Clusters - groups of atoms or molecules held together in atomically precise geometries by chemical interactions - offer significant potential for creating new materials with tailored properties across diverse applications \cite{eberhardtClustersNewMaterials2002}. Their unique characteristics distinguish them from both isolated molecules and bulk materials, presenting unique opportunities and challenges in computational modeling. Simulating inorganic clusters, particularly larger ones, is computationally demanding due to the necessity of accurate quantum mechanical calculations. This challenge is exacerbated for clusters containing heavy atoms like bismuth, where relativistic effects become significant and must be accounted for, further increasing computational complexity.

While ML-potentials have demonstrated the ability to predict energies and forces with high accuracy, most developments have focused on organic molecules and periodic materials. Organic molecules benefit from extensive datasets such as MD17 \cite{md17}, which facilitate the training of ML models. In contrast, datasets for inorganic clusters are virtually non-existent, making it difficult to train ML models or transfer knowledge from existing models.

To address this gap, we employ the PAL workflow to investigate the reactivity and transformations of small bismuth clusters (see Fig.~\ref{fig:app}c). Our goal is to demonstrate how ML-accelerated simulations can enhance our understanding of inorganic cluster formation and reactivity. In this application, we utilize graph neural networks as the \textbf{prediction kernel}, employing models like SchNet \cite{schutt_schnet_2018} and MACE \cite{Batatia2022mace}, which were originally designed for organic molecules.

We begin by pre-training these ML models on a foundational dataset of bismuth clusters to establish a baseline understanding of their behavior. Through the active learning workflow facilitated by PAL, we iteratively retrain the ML potentials on new configurations encountered during molecular dynamics (MD) simulations. The \textbf{generator kernel} produces MD trajectories for bismuth clusters of varying shapes, sizes, and charge states, effectively exploring the configurational space.

One significant challenge in modeling inorganic clusters is accounting for different charge states, as clusters can possess varying total charges leading to multiple potential energy surfaces. Reduction and oxidation processes involve interactions with environmental molecules, which are not explicitly modeled in this study. To address this, PAL's \textbf{oracle kernel} selectively labels highly uncertain and informative configurations using quantum mechanical calculations performed by Turbomole \cite{ahlrichs_electronic_1989}. These calculations provide accurate energies and forces for configurations that are poorly represented in the current training set.

The \textbf{training kernel} then incorporates these new data points to retrain the ML models, improving their accuracy and generalization to different charge states and cluster configurations. By dynamically updating the models, PAL enables them to capture complex information specific to bismuth cluster formation and their potential energy surfaces, extending beyond their original design focused on organic systems.

Similar to the first application, the bottleneck here is the labeling process in the oracle kernel, specifically for larger clusters. Additionally, as the ML model is not system-specific anymore, more complex ML potentials are used here, requiring flexibility and modularity to compare different ML potentials and potentially even move to periodic systems with explicit solvent molecules and counterions to even model redox-reaction events. Through the flexibility of PAL in combining different oracle and ML models, we effectively expand the scope of ML potentials to address the complexities inherent in inorganic clusters. This approach opens new avenues for simulating and understanding inorganic clusters, facilitating the development of materials with tailored properties.
\subsection{Thermo-fluid flow properties optimization}
As a final example of applying PAL, we move beyond the domain of atomistic modeling to demonstrate that the active learning workflows implemented in PAL are not limited to machine-learned potentials. Heat transfer in fluids is a complex process influenced by various factors such as geometry, material properties, and environmental conditions. While simple heat transfer problems can be solved analytically, most real-world scenarios require numerical simulations for accurate predictions. Computational fluid dynamics (CFD) is the primary method for numerically solving the governing equations of fluid mechanics and studying complex fluid flows without the need for physical experiments. However, high-fidelity CFD simulations are computationally expensive and time-consuming, despite providing high spatial and temporal resolution. This computational cost limits the ability to perform extensive parametric studies or real-time simulations.

To mitigate the high computational cost, machine learning models can be employed as surrogate models for CFD simulations, significantly reducing the overall computational cost while maintaining acceptable accuracy \cite{fukami2019super, sabater2022fast, vinuesa2021potential, duraisamy2021perspectives}. In our application, the primary objective is to predict thermo-fluid flow properties - specifically, the drag coefficient ($C_f$) and the Stanton number ($St$) - for two-dimensional laminar channel flows. Developing machine learning models capable of accurately predicting these fluid properties necessitates training datasets that encompass a wide variety of geometries and flow patterns. However, assembling such comprehensive datasets is challenging due to the computational expense of generating high-fidelity simulations for each configuration.

To address this challenge, we utilize PAL to strategically generate training data, thereby reducing the computational burden associated with simulating channel flows and leading to more efficient machine learning model development (see Fig.~\ref{fig:app}d). In this application of PAL: The \textbf{prediction kernel} consists of highly robust convolutional neural networks (CNNs) that are entirely invariant to image flipping and substantial shifting \cite{invariantcnn2024}. These CNNs serve as surrogate models that predict $C_f$ and $St$ from input geometries, enabling rapid evaluations without the need for full CFD simulations. The \textbf{generator kernel} employs particle swarm optimization (PSO) \cite{kennedy1995particle} to optimize the distribution of eddy-promoters within the flow domain \cite{kaithakkal2023heat}. Eddy-promoters are geometric features introduced to enhance mixing and heat transfer. By optimizing their placement, we can explore a diverse set of flow configurations that are most informative for training the surrogate models. The \textbf{oracle kernel} utilizes an in-house developed OpenFOAM solver \cite{mueller2021} to perform high-fidelity CFD simulations. These simulations compute the flow and temperature fields, as well as the corresponding fluid flow properties $C_f$ and $St$, providing accurate labels for the training data. The \textbf{training kernel} retrains the CNN models with the newly generated and labeled data, improving their predictive accuracy and generalization to new configurations. This iterative retraining ensures that the surrogate models remain accurate as they are exposed to new geometries and flow patterns.

Similar to the examples before, but in a very different application domain, there is not a unique bottleneck in any of the kernels but all kernels have similar computational costs. By integrating and parallelizing these components with PAL, we efficiently generate and select the most informative data for optimization while labeling them and training the ML surrogate model. Having the optimization process included in the generator allows us to focus computational resources on simulations that provide the greatest benefit not only for model performance but specifically for channel optimization, thereby reducing the total number of CFD simulations required to find good local or even the global optimum. As a result, we achieve surrogate models that can predict thermo-fluid flow properties of relevant close-to-optimal channel geometries with high accuracy while drastically reducing computational time compared to performing CFD simulations for every new configuration. This demonstrates PAL's versatility and effectiveness \textit{beyond atomistic simulations}, highlighting its potential for applications in engineering domains where computational cost is a limiting factor. By leveraging PAL, we can perform rapid optimization and design studies in thermo-fluid systems, contributing to advancements in fields such as heat exchanger design, microfluidics, and aerodynamic surface optimization.

\begin{table}[h]
\centering
\begin{tabular}{p{2.5cm} | p{3.5cm} | p{3.5cm} | p{3.5cm}}
\hline
\textbf{Application} & \textbf{Prediction \& training kernel} & \textbf{Generator kernel} & \textbf{Oracle kernel} \\
\hline
Photodynamics simulations & Fully connected neural network committee & Parallel surface-hopping MD simulations & TDDFT (B3LYP/6-31G*) with Turbomole \\
\hline
HAT simulations & Graph neural network (SchNet, Allegro, MACE) committee & Randomized sampling of relevant geometries; transition state search & Semiempirical calculations with xTB and DFT (BMK/def2-TZVPD) with Turbomole \\
\hline
Inorganic clusters & Graph neural network (SchNet, MACE) committee & MD simulations with varying cluster sizes and charges & DFT (TPSS/dhf-TZVP) with Turbomole \\
\hline
Thermo-fluid flow optimization & Convolutional neural network committee & Particle Swarm Optimization & CFD simulations: In-house OpenFOAM solver \\
\hline
\end{tabular}
\caption{Summary of Applications and Corresponding Kernel Choices in PAL}
\label{tab:applications}
\end{table}


\section{Discussion: Library use, current limitation, and future developments}
\textbf{Using PAL to customize and automate active learning workflows.} With increasing popularity of ML potentials, not only in proof-of-principle studies but also in exploratory research of novel systems, also the role of active learning increases, and thus the need for efficient, customizable, and easy-to-use implementations. While different application scenarios have different requirements on the specific methods used in active learning, i.e. the ML models and the data generation algorithms and methods, the general workflow and the required communication backbone is the same in almost all cases. Thus, this backbone does not need to be implemented by every researcher or research group. Furthermore, in the spirit of open software, specific parts and modules in active learning workflows should be shared by users to make it easier in the future to build new active learning workflows. Our parallel active learning library PAL is expected to be used by researchers to integrate, customize, and develop different parts of active learning workflows without having to reimplement the entire communication and workflow scheduling backbone again and again. PAL helps to automate active learning workflow while minimizing the effort of modifications in the implementation. Thus, we believe it can help significantly enhance the efficiency of constructing training sets, training ML potentials, and applying them to interesting systems.

The current source code of PAL includes the general and generic backbone for communication and workflow automation as well as blueprints/placeholders of the different application-specific kernels - ML potential training and prediction, oracle, and generator. We furthermore provide example implementations of those modules for the four application examples discussed in Section~\ref{fig:app}. Those can easily be used, mixed, and adapted by users. We plan to develop more prototypical kernels in the future which will make it easier for users to combine them to active learning workflows with miminal coding effort. We also encourage users to contribute additional kernels to the library, to make them accessible to other researchers.

\textbf{Hardware.}. The current PAL workflow is implemented and tested on the Slurm workload manager, using a single type of computational node (CPU node, GPU node, or CPU-GPU hybrid node). Executing PAL on other scheduling systems will require additional user input to specify computational resources. We plan to extend this work to support additional batch systems and enhance flexibility in node scheduling, job assignment, and resource management. Another future goal is to incorporate real-time tracking and monitoring of timing and resource usage, such as GPU and memory utilization, to facilitate workflow optimization in various scenarios. 

\textbf{Communictation bottleneck.} When the inference time of ML models in the prediction kernel is $10 \mathrm{ms}$ or less, communication between the generator and predictor can become a bottleneck limiting the speed of exploration in the generator. Solving this requires further work to couple generator and prediction kernel more tightly, though this is not currently an issue for typical ML potentials, especially complex models such as equivariant graph neural networks. Additionally, if the shape of the ML input or output is not fixed, there is an overhead as MPI messages require a predetermined size to be efficient.

\textbf{Available kernels.} As discussed in Section~\ref{fig:app} and above, the PAL library currently includes a few examples of generators, ML models, and oracles, primarily in the fields of materials science, chemistry, and engineering. We plan to expand these examples to encompass more relevant applications of active learning in the future.

\section{Conclusion}
In this work, we introduced PAL, an automated, modular, and parallel active learning library designed to overcome the limitations of traditional active learning workflows that often require extensive human intervention and underutilize modern computational resources. By decomposing the active learning process into five modular core kernels, the prediction kernel, generator kernel, training kernel, oracle kernel, and controller kernel, PAL enables asynchronous and parallel execution of data generation, labeling, model training, and prediction tasks. This modular architecture allows users to easily customize and extend each component to suit a wide range of applications across different scientific and engineering domains.

Our examples illustrate how PAL significantly reduces computational overhead and improves scalability, achieving substantial speed-ups through asynchronous parallelization on both CPU and GPU hardware. By decoupling the prediction and training processes, PAL minimizes disruptions caused by time-consuming labeling and model updates, ensuring efficient utilization of high-performance computing resources. The library's flexibility and effectiveness are showcased through its successful application to diverse real-world scenarios, including photodynamics simulations of organic semiconductors, hydrogen atom transfer reactions in biological systems, atomistic simulations of inorganic clusters, and thermo-fluid flow optimization in engineered systems.

PAL advances the field of scientific active learning by providing a scalable and adaptable framework that streamlines the integration of machine learning models, uncertainty estimation methods, oracles, and data exploration strategies. It facilitates the development of highly accurate models with minimal data acquisition costs, thereby accelerating research and discovery in various domains. The ability to handle complex workflows and large-scale computations makes PAL a valuable tool for scientists and engineers seeking to leverage active learning in their work.

Looking ahead, future developments of PAL will focus on enhancing its capabilities and user experience. Plans include supporting additional batch systems HPC environments, incorporating real-time monitoring and resource management features, and integrating with other machine learning frameworks and tools. We also aim to expand the library's documentation and provide comprehensive tutorials to lower the adoption barrier for new users. By addressing current limitations and fostering community contributions, we hope that PAL becomes a useful and widely applied tool in active learning workflows of ML potentials and beyond, empowering researchers to efficiently harness computational resources and drive innovation in their respective fields.

\subsection*{Code availability}
The data and models supporting the application use cases discussed here can be found in the individual publications. The code for PAL is available through the open-source repository on GitHub for continuous development, also by the community: \url{https://github.com/aimat-lab/PAL}. 


\subsection*{Author contributions}
P.F. and C.Z. contributed to the conceptualization of the project. C.Z. is the main developer of the methodology and software. M.N., Y.K., and Y.Z. contributed aspects of the software. P.F. contributed resources, supervision, and funding acquisition. All authors contributed to writing the origianl draft as well as to visualization of the content.

\subsection*{Conflicts of interest}
There are no conflicts of interest to declare.

\subsection*{Acknowledgements}
P.F. acknowledges support by the Federal Ministry of Education and Research (BMBF) under Grant No. 01DM21001B (German-Canadian Materials Acceleration Center).
M.N. acknowledges funding from the Klaus Tschira Stiftung gGmbH (SIMPLAIX project 1).
Y.K. acknowledges funding by the German Research Foundation (Deutsche Forschungsgemeinschaft, DFG) within Priority Programme SPP 2331.
The project is co-funded by the European Union (ERC, BiCMat, 101054577). Views and opinions expressed are however those of the authors only and do not necessarily reflect those of the European Union or the European Research Council. Neither the European Union nor the granting authority can be held responsible for them.
This work was performed on the HoreKa supercomputer funded by the Ministry of Science, Research and the Arts Baden-Württemberg and by the Federal Ministry of Education and Research.
The authors acknowledge support by the state of Baden-Württemberg through bwHPC.

\printnomenclature

\clearpage



\bibliography{bib}
\bibliographystyle{rsc} 

\clearpage

\section*{Supporting information}
\renewcommand{\thesection}{S\arabic{section}}
\renewcommand{\thesubsection}{\thesection.\arabic{subsection}}
\setcounter{section}{0} 
\setlength{\parindent}{0pt}

\section{Parallelization with MPI}
\begin{figure}
  \centering
  \includegraphics[width=\textwidth]{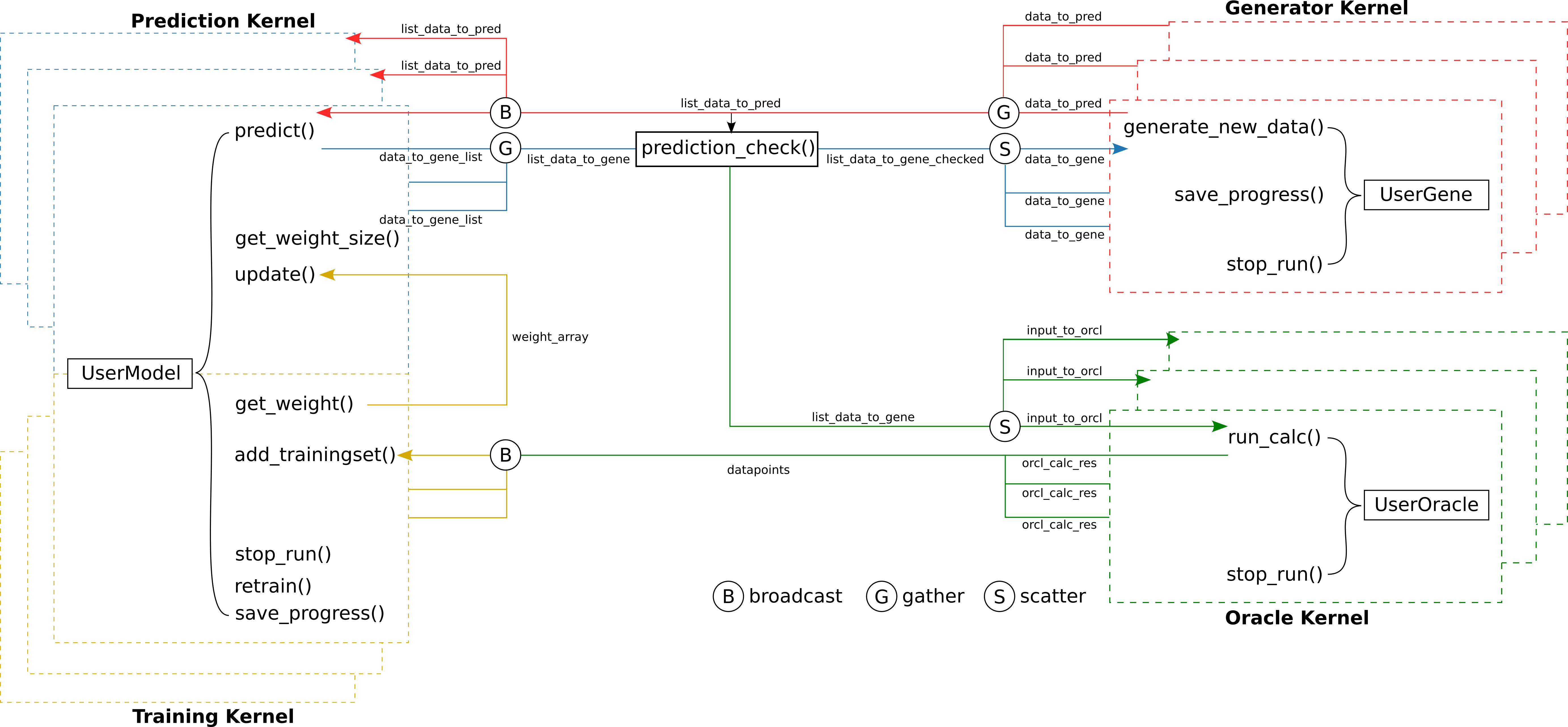}
  \caption{
  Methods and data flow of PAL.
  }
  \label{fig:funcs}
\end{figure}

The PAL is designed in an object-oriented manner, wherein the implementation of each kernel is for a single corresponding process and the user-specified number of processes are generated for every kernel during execution. The methods of different processes and data flow among them are displayed in Figure~\ref{fig:funcs}. Each process is assigned a unique ID (MPI rank) and maintains its data, states, and behaviors. For communication efficiency, data transferred among kernels should be arranged as 1-D Numpy numerical arrays. The communication processes are handled by the Controller kernel separated from the user with MPI message-passing methods, including broadcast, gather, scatter, and point-to-point communications. Explanations of every kernel are presented in the following sections with a toy example in which random numbers are generated in Generator and Oracle kernels and sent to Prediction and Training kernels for inference and training.

\section{Speedup estimation for parallel active learning workflow}
\label{appendix:speedup_estimation}

In this section, we present a detailed estimation of the speedup achievable by the Parallel Active Learning (PAL) workflow compared to a fully serial active learning workflow. We define the relevant parameters, derive the runtime equations for both workflows, and provide speedup estimates for specific use cases.

\subsection{Relevant parameters, runtimes, and speedup}

To estimate the runtime and speedup, we consider the following parameters in a simplified setting:

\begin{itemize}
    \item $t_{\text{oracle}}$: Time to label a single sample using the oracle.
    \item $t_{\text{train}}$: Time to train the machine learning (ML) model.
    \item $t_{\text{gen}}$: Time to perform 1000 steps of the generator and predictor modules.
    \item $N$: Number of samples to be labeled.
    \item $P$: Number of parallel workers available for labeling. We always assume that $P \leq N$
\end{itemize}

\textbf{Rumtime of the serial workflow}

In the serial active learning workflow, assuming only parallelization of the oracles, the processes are executed sequentially in the following order: labeling, training, generation/prediction. The total runtime for $N$ samples is given by:

\begin{equation}
    T_{\text{serial}} = \frac{N}{P} \cdot t_{\text{oracle}} + t_{\text{train}} + t_{\text{gen}}
\end{equation}\\

\textbf{Rumtime of the parallel workflow}

In the PAL workflow, the labeling, training, generation, and prediction modules operate in parallel, coordinated by the central module. Assuming perfect parallelization and ignoring communication overhead, the runtime can be approximated by the maximum of the individual module runtimes:

\begin{equation}
    T_{\text{parallel}} = \max\left(\frac{N}{P}\cdot t_{\text{oracle}}, t_{\text{train}}, t_{\text{gen}}\right)
\end{equation}\\

\textbf{Speedup}

The speedup ($S$) achieved by the PAL workflow over the serial workflow is defined as:

\begin{equation}
    S = \frac{T_{\text{serial}}}{T_{\text{parallel}}}
\end{equation}

Substituting the expressions for $T_{\text{serial}}$ and $T_{\text{parallel}}$, we obtain:

\begin{equation}
    S = \frac{\frac{N}{P} \cdot t_{\text{oracle}} + t_{\text{train}} + t_{\text{gen}}}{\max\left(\frac{N}{P}\cdot t_{\text{oracle}}, t_{\text{train}}, t_{\text{gen}}\right)}
\end{equation}

Note that this speedup is only a lower bound of the speedup, as in the parallel workflow, all available resources are used and never idle. Thus, if the oracle is the bottleneck, the training and the generation/prediction tasks are continued and thus train over more epochs and explore more parts of the input space than the serial version of the workflow.

\subsection{Use case specific estimates}

We consider three distinct use cases with varying oracle methods, ML models, and generators. For each use case, we estimate the speedup based on the provided parameters.\\

\textbf{Use Case 1: DFT and GNNs}

\begin{itemize}
    \item Oracle: DFT, $t_{\text{oracle}} \approx 1$ hour
    \item ML Model: Graph Neural Network (GNN), $t_{\text{train}} \approx 1$ hour
    \item Generator: Molecular Dynamics (MD), $t_{\text{gen}} \ll 1$ hour
\end{itemize}

Assuming $P$ parallel workers:

\begin{align}
    T_{\text{serial}} &= \frac{N}{P} \cdot t_{\text{oracle}} + t_{\text{train}} + t_{\text{gen}} \approx \frac{N}{P} \cdot t_{\text{oracle}} + t_{\text{train}} \\
    T_{\text{parallel}} &= \max\left(\frac{N}{P} \cdot t_{\text{oracle}}, t_{\text{train}}, t_{\text{train}}\right) = \max\left(\frac{N}{P}\cdot t_{\text{oracle}}, t_{\text{train}}\right)
\end{align}

Thus, the speedup is:

In case that oracle evaluations and model training require similar computation times ($t_{\text{oracle}} = t_{\text{train}} = t$), and if we assume $N \geq P$ (leading to $T_{\text{parallel}} = \frac{N}{P} \cdot t_{\text{oracle}}$), the speedup obtained through PAL is

\begin{equation}
    S = \frac{\frac{N}{P}\cdot t + t}{\frac{N}{P}\cdot t}
      = \frac{\frac{N}{P} + 1}{\frac{N}{P}} = \left(1 + \frac{P}{N}\right)
\end{equation}

This means that in the case of balanced costs between oracle and training modules, the highest speedup is obtained if the number of parallel workers for labeling $P$ is equal to the number of oracle calls $N$ per iteration. This ratio is automatically optimized by PAL through asynchronous communication, leading to a speedup of 2 compared to serial active learning workflows.

However, if DFT calculations are much more time-consuming than model training, parallelization through a large number of parallel oracle calculations $P$ is the limiting factor, indicating that best results are achieved if $P$ in maximized.\\

\textbf{Use Case 2: Semiempirical oracle for reaction network exploration}

\begin{itemize}
    \item Oracle: xTB, $t_{\text{oracle}} = 10$ seconds
    \item ML Model: Graph Neural Network (GNN), $t_{\text{train}} = 1$ hour
    \item Generator: Transition state earch, $t_{\text{gen}} = 10$ minutes
\end{itemize}

In this case, the bottleneck is clearly the model training, as during training time, $360P$ new calculations can be done by the oracle. With a clear bottleneck, the parallel and serial runtime are approximately the same, leading to no substantial speedup through parallelization:

\begin{align}
    T_{\text{serial}} &\approx t_{\text{train}}\\
    T_{\text{parallel}} &\approx \max\left(t_{\text{oracle}} + t_{\text{train}} + t_{\text{gen}}\right) = t_{\text{train}}\\
    S &\approx \frac{t_{\text{train}}}{t_{\text{train}}} = 1
\end{align}

In this particular case, we recommend using a rolling training set where newly incoming xTB-labeled samples are added after every single training epoch, and old samples are removed to keep the training set size constant. This limits the training time per epoch and at the same time avoids overfitting to a fixed training set, ensuring permanent domain adaptation to the relevant molecules and geometries in the region of the chemical space that is currently explored by the generator. This dynamic and asynchronous way how data is communicated between modules in PAL makes this easily possible without users to write dedicated code for communication and scheduling.\\

\textbf{Use Case 3: Computational fluid dynamics (CFD)}

\begin{itemize}
    \item Oracle: CFD, $t_{\text{oracle}} = 10$ minutes
    \item ML Model: Convolutional Neural Network (CNN), $t_{\text{train}} = 10$ minutes
    \item Generator: Particle Swarm Optimization (PSO), $t_{\text{gen}} = 10$ minutes
\end{itemize}

In this scenario, assuming $P = N$ parallel workers, and defining $t_{\text{oracle}} = t_{\text{train}} = t_{\text{gen}} = t$, all three modules have balanced costs, leading to the following runtimes:

\begin{align}
    T_{\text{serial}} &= t_{\text{oracle}} + t_{\text{train}} + t_{\text{gen}} = 3t\\   T_{\text{parallel}} &= \max\left(t_{\text{oracle}} + t_{\text{train}} + t_{\text{gen}}\right) = t
\end{align}

Thus, the speedup is:

\begin{equation}
    S = \frac{3t}{t} = 3
\end{equation}

Thus, in scenarios without a clear bottleneck, i.e. where the computational cost of all modules is similar, the speedup approaches 3, independent of other considerations. Thus, when designing parallel active learning workflows with PAL, it is desirable to break down the computational steps in all modules into small pieces (one single oracle call, potentially with sub-parallelization, one single epoch of training, and one short generator run to explore the input space), allowing load-balancing and thus optimal parallelization for high speedup factors.

\subsection{Conclusion}

The Parallel Active Learning workflow presents a substantial improvement over traditional serial approaches by effectively utilizing high-performance computing resources. The speedup estimations for various use cases highlight PAL's versatility and efficiency, making it a valuable tool for researchers in computational science and beyond.

\section{PAL settings}\label{sec:setting}
Settings of a toy PAL example are displayed in the code block below. The user defines the number of processes in each kernel by setting \texttt{pred\_process}, \texttt{orcl\_process}, \texttt{gene\_process}, and \texttt{ml\_process}. The total number of processes initialized by PAL should be the summation of processes in the four kernels with two additional processes for the Controller. The user can also specify the distribution of processes among computational nodes by setting \texttt{designate\_task\_number} as \texttt{True} and designating the distribution in \texttt{task\_per\_node}. As the MPI requires to know the message size at the receiving end before communication, \texttt{fixed\_size\_data} should be set to \texttt{False} if inputs from Generator processes or predictions from Prediction processes vary in size for different data points, in which case sizes of data are passed first for every MPI communication.

\begin{lstlisting}[language=Python]
AL_SETTING = {
    "result_dir": '../results/TestRun',    # directory to save all metadata and results
    "orcl_buffer_path": '../results/TestRun/ml_buffer',    # path to save data ready to send to ML. Set to None to skip buffer backup.
    "ml_buffer_path": '../results/TestRun/orcl_buffer',    # path to save data ready to send to Oracle. Set to None to skip buffer backup.

    # Number of process in total = 2 MPI communication processes (Manager and Exchange)
    #                              + pred_process + orcl_process + gene_process + ml_process
    "pred_process": 3,                     # number of prediction processes
    "orcl_process": 5,                     # number of oracle processes
    "gene_process": 20,                    # number of generator processes
    "ml_process": 3,                       # number of machine learning processes
    "designate_task_number": True,         # set to True if need to specify the number of tasks running on each node (e.g. number of model per computation node)
                                           # if False, tasks are arranged randomly
    "fixed_size_data": True,              # set to True if data communicated among kernels have fixed sizes.
                                           # if false, additional communications are necessary for each iteration to exchange data size info thus lower efficiency.
    "task_per_node":{                      # designate the number of tasks per node, used only if designate_task_number is True
        "prediction": [3, 0],              # list for the number of tasks per node (length must matches the number of nodes), None for no limit
        "generator": None,                 # list for the number of tasks per node (length must matches the number of nodes), None for no limit
        "oracle": None,                    # list for the number of tasks per node (length must matches the number of nodes), None for no limit
        "learning": [0, 3],                # list for the number of tasks per node (length must matches the number of nodes), None for no limit
    },
    "orcl_time": 10,                       # Oracle calculation time in seconds
    "progress_save_interval": 60,          # time interval (in seconds) to save the progress
    "retrain_size": 20,                    # batch size of increment retraining set
    "dynamic_orcale_list": True,           # adjust data points for orcale calculation based on ML predictions everytime when retrainings finish
    "gpu_pred": [],                        # gpu index list for prediction processes
    "gpu_ml": [],                          # gpu index list for machine learning
    "usr_pkg": {                           # dictionary of paths to user implemented modules (generator, model, oracle and utils)
        "generator": "../usr_example/generator.py",
        "model": "../usr_example/model.py",
        "oracle": "../usr_example/oracle.py",
        "utils": "../usr_example/utils.py",
    },
    }
\end{lstlisting}

\section{Prediction kernel}
This kernel is implemented together with the Training kernel and behaviors differently from the latter by calling different methods during execution. The initialization of a prediction process is shown in the code below where the process ID (PID or rank), path to the working directory (\texttt{result\_dir}), and device index (\texttt{i\_device}) are assigned by the Controller. 

\begin{lstlisting}[language=Python]
def __init__(self, rank, result_dir, i_device, mode):
    """
    Initilize the model.
        
    Args:
        rank (int): current process rank (PID).
        result_dir (str): path to directory to save metadata and results.
        i_device (int): Index for assigning computation to device (GPU or CPU).
        mode (str): 'predict' for Prediction and 'train' for Machine Learning.
    """
    self.rank = rank
    self.result_dir = result_dir
    self.mode = mode
    self.i_device = i_device
        
    if self.mode == "predict":
        self.model = TestModel(4, 4)
        self.para_keys = list(self.model.state_dict().keys())

    else:
        ... ...
\end{lstlisting}

For inference, the Prediction process takes a list of 1-D arrays (\texttt{list\_data\_to\_pred}) that consists of output values (\texttt{data\_to\_pred}) of \texttt{generate\_new\_data} method of Generator processes gathered by the Controller (Figure~\ref{fig:funcs}, flow in red). In cases where multiple Prediction processes are running in parallel, the same copy of \texttt{list\_data\_to\_pred} is broadcasted to each process. The prediction method handles the organization of arguments, prediction of target values, and organization of values returning to the Generator kernel. The prediction method should return a list of 1-D arrays (\texttt{data\_to\_gene\_list}) that are distributed to each process in the Generator kernel, thus the size of \texttt{data\_to\_gene\_list} and the order of 1-D arrays should match processes in Generator kernel. The \texttt{data\_to\_gene\_list} returned by every Prediction process are gathered by the Controller and passed through the user-defined \texttt{prediction\_check} function before scattering to Generator processes' \texttt{generate\_new\_data} method (Figure~\ref{fig:funcs}, flow in blue). If the PAL workflow is shutdown by any Training process or Generator process, the \texttt{stop\_run} method is called by every Prediction process before quitting.

\begin{lstlisting}[language=Python]
def predict(self, list_data_to_pred):
    """
    Make prediction for list of inputs from Generator.
    
    Args:
        list_data_to_pred (list): list of user defined model inputs. [1-D numpy.ndarray, 1-D numpy.ndarray, ...]
                           size is equal to number of generator processes
                           Source: list of data_to_pred from UserModel.generate_new_data().
        
    Returns:
        data_to_gene_list (list): predictions returned to Generator. [1-D numpy.ndarray, 1-D numpy.ndarray, ...]
                                  size should be equal to number of generator processes
                                  Destination: list of data_to_gene at UserModel.generate_new_data().
    """
    data_to_gene_list = None
    
    ##### User Part #####
    # organize the input data into a ndarray
    input_array = np.array(list_data_to_pred, dtype=float)

    self.model.eval()
    with torch.no_grad():
        x = torch.tensor(input_array, dtype=torch.float)
        #print(f"Debug Rank {self.rank}: data_to_gene shape {x.shape}")
        data_to_gene_list = list(self.model(x).numpy())

    # data_to_gene_list should be a list containing 1-D numpy arrays
    return data_to_gene_list

def stop_run(self):
    """
    Called before the Prediction/Training process terminating when active learning workflow shuts down.
    """
    ##### User Part #####
    if self.mode == "train":
        self.save_progress()
\end{lstlisting}

The ML model in each Prediction process is updated by replicating weight parameters from the corresponding model in the Training kernel, where the weights are arranged as a 1-D array (\texttt{weight\_array}) by the Training process \texttt{get\_weight} method and sent to the Prediction process directly. The Prediction process then calls the \texttt{update} method to reorganize the weight array and update the ML model. As the MPI requires to know the message size at the receiving end before communication, the \texttt{get\_weight\_size} method is called once when PAL workflow is initialized to return the size of the 1-D \texttt{weight\_array}.

\begin{lstlisting}[language=Python]
def update(self, weight_array):
    """
    Update model/scalar with new weights in weight_array.
    
    Args:
        weight_array (1-D numpy.ndarray): 1-D numpy array containing model/scalar weights.
                                          Source: weight_array from UserModel.get_weight().
    """
    ##### User Part #####
    for k in self.para_keys:
        self.model.state_dict()[k] = weight_array[:self.model.state_dict()[k].flatten().shape[0]].reshape(self.model.state_dict()[k].shape)
    print(f"Prediction Rank {self.rank}: model updated")

def get_weight_size(self):
    """
    Return the size of model weight when unpacked as an 1-D numpy array.
    Used to send/receive weights through MPI.
    
    Returns:
        weight_size (int): size of model weight when unpacked as an 1-D numpy array.
    """
    weight_size = None
    
    ##### User Part #####
    weight_size = 0
    for k in self.para_keys:
        weight_size += self.model.state_dict()[k].flatten().shape[0]

    # weight_size should be returned as an integer
    return weight_size
\end{lstlisting}

\section{Training kernel}
The Training kernel is implemented together with the Prediction kernel and distinguished from the former by the \texttt{mode} flag. Other than training/retraining ML models, Training processes are also responsible for saving data of interest, such as training history and model parameters. This is achieved by the \texttt{save\_progress} method that is called whenever the training stops or restarts. 

\begin{lstlisting}[language=Python]
def __init__(self, rank, result_dir, i_device, mode):
    """
    Initilize the model.
    
    Args:
        rank (int): current process rank (PID).
        result_dir (str): path to directory to save metadata and results.
        i_device (int): Index for assigning computation to device (GPU or CPU).
        mode (str): 'predict' for Prediction and 'train' for Machine Learning.
    """
    self.rank = rank
    self.result_dir = result_dir
    self.mode = mode
    self.i_device = i_device
    
    if self.mode == "predict":
        ... ...
    
    else:
        self.start_time = time.time()
        self.x_train = []
        self.y_train = []
        self.x_val = []
        self.y_val = []
        self.val_split = 0.2
        self.model = TestModel(4, 4)
        self.para_keys = list(self.model.state_dict().keys())
        self.loss = nn.MSELoss(reduction='sum')
        self.optimizer = torch.optim.Adam(self.model.parameters(), lr=1e-3)
        self.max_epo = 1000000
        self.batch_size = 10
        self.history = {
            "MSE_train": [],
            "MSE_val": []
            }

def save_progress(self):
    """
    Save the current progress/data/state.
    Called everytime after retraining and receiving new data points.
    """
    ##### User Part #####
    with open(os.path.join(self.result_dir, f"retrain_history_{self.rank}.json"), 'w') as fh:
        json.dump(self.history, fh)
\end{lstlisting}

Once the user-defined number of new data are collected by the Controller from the Oracle kernel, the data collection is broadcasted to each Training process as a list of lists (\texttt{datapoints}). Each sub-list contains two 1-D arrays with the first one being the input and the second being label. The Prediction processes call \texttt{add\_trainingset} to organize the new data and add to the training/validation set (Figure~\ref{fig:funcs}, flow in yellow). 

\begin{lstlisting}[language=Python]
def add_trainingset(self, datapoints):
    """
    Increase the training set with set of data points.
    
    Args:
        datapoints (list): list of new training datapoints.
                           Format: [[input1 (1-D numpy.ndarray), target1 (1-D numpy.ndarray)], [input2 (1-D numpy.ndarray), target2 (1-D numpy.ndarray)], ...]
                           Source: input_for_orcl element of input_to_orcl_list from utils.prediction_check(). 
                                   orcl_calc_res from UserModel.run_calc().
    """
    ##### User Part #####
    idx = np.array(range(0, len(datapoints)), dtype=int)
    val_size = int(self.val_split * idx.shape[0])
    i_val = np.random.choice(idx, size=val_size, replace=False)
    i_train = np.array([i for i in idx if i not in i_val], dtype=int)
    for i in range(0, len(datapoints)):
        if i in i_train:
            self.x_train.append(datapoints[i][0])
            self.y_train.append(datapoints[i][1])
        else:
            self.x_val.append(datapoints[i][0])
            self.y_val.append(datapoints[i][1])
    print(f"Training Rank{self.rank}: training set size increased")
\end{lstlisting}

The Training processes keep training/retraining ML models, which can be halted by a): new training data arriving or b): a user-defined early stopping criterion met, where a) is achieved by checking \texttt{req\_data.Test()} at every training epoch and stops when the return value is \texttt{True}. The entire PAL workflow can be shut down by any one of the Training processes if the return flag \texttt{stop\_run} is set to \texttt{True}, in which scenario each Training process calls \texttt{stop\_run} method before quitting. Otherwise, the process is halted until new training data arrive and the retraining restarts by repeatedly calling the \texttt{retrain} method.

\begin{lstlisting}[language=Python]
def retrain(self, req_data):
    """
    Retrain the model with current training set.
    Retraining should stop before or when receiving new data points.
    
    Args:
        req_data (MPI.Request): MPI request object indicating status of receiving new data points.

    Returns:
        stop_run (bool): flag to stop the active learning workflow. True for stop.
    """
    stop_run = False
    ##### User Part #####
    print(f"Training Rank{self.rank}: retraining start...")
    for v in self.history.values():
        v.append([])
    x_t = torch.tensor(np.array(self.x_train, dtype=float), dtype=torch.float)
    y_t = torch.tensor(np.array(self.y_train, dtype=float), dtype=torch.float)
    x_v = torch.tensor(np.array(self.x_val, dtype=float), dtype=torch.float)
    y_v = torch.tensor(np.array(self.y_val, dtype=float), dtype=torch.float)
    n_batch = int(len(self.x_train) / self.batch_size)
    n_batch_val = int(len(self.x_val) / self.batch_size)
    for i in range(1, self.max_epo+1):
        self.model.train()
        mse = 0
        for j in range(1, n_batch+1):
            pred = self.model(x_t[j*self.batch_size:min((j+1)*self.batch_size, x_t.shape[0])])
            loss = self.loss(pred, y_t[j*self.batch_size:min((j+1)*self.batch_size, y_t.shape[0])])
            
            loss.backward()
            self.optimizer.step()
            self.optimizer.zero_grad()
            mse += loss.item()
        self.history["MSE_train"][-1].append(mse/x_t.shape[0])
        if i % 10 == 0:
            self.model.eval()
            mse = 0
            for j in range(1, n_batch_val+1):
                with torch.no_grad():
                    pred = self.model(x_v[j*self.batch_size:min((j+1)*self.batch_size, x_v.shape[0])])
                    loss = self.loss(pred, y_v[j*self.batch_size:min((j+1)*self.batch_size, y_v.shape[0])]).item()
                    mse += loss
            self.history["MSE_val"][-1].append(mse/x_v.shape[0])
            
        # req_data.Test() indicates if new data have arrived from Oracle through MG
        if req_data.Test():
            # if new data arrive, stop retraining to update training/validation set
            print(f"Training Rank{self.rank}: new data arrive.")
            break
    print(f"Training Rank{self.rank}: retraining stop.")
    if time.time() - self.start_time >= 3600:
        stop_run = True
        print(f"Training Rank{self.rank}: stop signal sent.")
    else:
        stop_run = False
    
    # stop_run should be returned as a bool value.
    return stop_run

def stop_run(self):
    """
    Called before the Prediction/Training process terminating when active learning workflow shuts down.
    """
    ##### User Part #####
    if self.mode == "train":
        self.save_progress()
\end{lstlisting}

\section{Generator kernel}
Processes in the Generator kernel are generated with its unique rank and working directory (\texttt{result\_dir}). Each Generator process keeps its own records of state and data such as previously generated data or the number of inference-generation iterations, and saves records to \texttt{result\_dir} through \texttt{save\_progress} method after a user-defined interval (\texttt{progress\_save\_interval} in \texttt{al\_setting}). Processes in the Generator kernel are alive through the entire PAL lifetime. Killing and restarting a Generator process is impossible in current PAL, while similar behavior could be achieved in the \texttt{generate\_new\_data} method.

\begin{lstlisting}[language=Python]
def __init__(self, rank, result_dir):
    """
    initilize the generator.
    
    Args:
        rank (int): current process rank (PID).
        result_dir (str): path to directory to save metadata and results.
    """
    self.rank = rank
    self.result_dir = result_dir
    ##### User Part ######
    self.counter = 0
    #self.limit = float("inf")
    self.limit = 300000 + self.rank
    self.state = np.random.randn(4)
    self.history = [[],]
    self.save_path = os.path.join(self.result_dir, f"generator_data_{rank}")

def save_progress(self):
    """
    Save the current state and progress. Called everytime after the interval defined by progress_save_interval in al_setting.
    """
    ##### User Part #####
    m = 'ab' if os.path.exists(self.save_path) else 'wb'
    with open(self.save_path, m) as fh:
        pickle.dump(self.history[:-1], fh)
    self.history = self.history[-1:]
\end{lstlisting}

The Generator processes call \texttt{generate\_new\_data} method to generate new data and organize them as 1-D arrays (\texttt{data\_to\_pred}) that are gathered by Controller and broadcasted to Prediction processes for inference (Figure~\ref{fig:funcs}, flow in red).  The \texttt{generate\_new\_data} method takes \texttt{data\_to\_gene} as input, which contains \texttt{None} value at the first AL iteration and predictions from the Prediction kernel through the \texttt{prediction\_check} function for the rest. The entire PAL workflow can be shut down by any one of the Generator processes if the return flag \texttt{stop\_run} is set to \texttt{True}, in which scenario each Generator process calls \texttt{stop\_run} method before quitting.

\begin{lstlisting}[language=Python]
def generate_new_data(self, data_to_gene):
    """
    Generate new data point based on data_to_gene (prediction from Prediction kernel).
    
    Args:
        data_to_gene (1-D numpy.ndarray or None): data from prediction kernel through EXCHANGE process.
                                                  Initialized as None for the first time step.
                                                  Source: element of data_to_gene_list from UserModel.predict()
        
    Returns:
        stop_run (bool): flag to stop the active learning workflow. True for stop.
        data_to_pred (1-D numpy.ndarray): data to prediction kernel through EXCHANGE process.
                                          Destination: element of input_list at UserModel.predict()
    """
    stop_run = False
    data_to_pred = None
    
    # please notice that data_to_gene is intinilized to be None for the first iteration.
    ##### User Part #####
    # in this simple example, generator processes return random numbers
    if self.counter > self.limit:
        stop_run = True
        data_to_pred = np.random.randn(4)
        print(f"Generator rank{self.rank}: stop signal sent.")
    elif data_to_gene is None:
        data_to_pred = np.random.randn(4)
        self.history.append([data_to_pred,])
    elif (data_to_gene == 0).any():
        data_to_pred = np.random.randn(4)
        self.history.append([data_to_pred,])
    else:
        data_to_pred = self.state * data_to_gene
        self.history[-1].append(data_to_pred)  

    if self.counter % 10000 == 0:
        print(f"Generator rank{self.rank}: iteration {self.counter} finished.")    #TODO remove debug later
    
    self.counter += 1
    
    # stop_run should be returned as a bool value
    # data_to_pred should be returned as an 1-D numpy array
    return stop_run, data_to_pred

def stop_run(self):
    """
    Called before the Generator process terminating when active learning workflow shuts down.
    """
    ##### User Part #####
    self.save_progress()
\end{lstlisting}

\section{Oracle kernel}
Each process in the Oracle kernel is started with an MPI rank and a working directory (\texttt{result\_dir}). Once the PAL is shutdown by any Training process or Generator process, the \texttt{stop\_run} method is called by every Oracle process before quitting.

\begin{lstlisting}[language=Python]
def __init__(self, rank, result_dir):
    """
    Initilize the model.
    
    Args:
        rank (int): current process rank (PID).
        result_dir (str): path to directory to save metadata and results.
    """
    self.rank = rank
    self.result_dir = result_dir

def stop_run(self):
    """
    Called before the Oracle process terminating when active learning workflow shuts down.
    """
    ##### User Part #####
    pass
\end{lstlisting}

Oracle processes generate ground-truth labels for ML model training by repeatedly calling the \texttt{run\_calc} method with a 1-D arry (\texttt{input\_for\_orcl}) as input, which is selected by the \texttt{prediction\_check} function (\text{list\_input\_to\_orcl}). The input data (\texttt{input\_for\_orcl}) and ground-truth label (\texttt{orcl\_calc\_res}) are then collected by Controller before broadcasting to Training processes (Figure~\ref{fig:funcs}, flow in green).

\begin{lstlisting}[language=Python]
def run_calc(self, input_for_orcl):
    """
    Run Oracle computation.
    Args:
        input_for_orcl (1-D numpy.ndarray): input for oracle computation.
                                            Source: element of input_to_orcl_list from utils.prediction_check()

    Returns:
        orcl_calc_res (1-D numpy.ndarray): results generated by Oracle.
                                           Destination: element of datapoints at UserModel.add_trainingset().
    """
    orcl_calc_res = None
    ##### User Part #####
    orcl_calc_res = np.random.randn(4,)
    
    # orcl_calc_res should be returned as an 1-D numpy array
    return orcl_calc_res
\end{lstlisting}

\subsection*{Utilities}
The \texttt{prediction\_check} function takes as arguments a list of predictions gathered from the Prediction kernel and a list of inputs gathered from the Generator kernel, and selects inputs that are sent to the Oracle kernel for labeling by user-defined criteria (\texttt{list\_input\_to\_orcl}, e.g. standard deviation of ensemble model predictions). Data returned as \texttt{list\_data\_to\_gene\_checked} are scattered to Generator processes for the next AL iteration. Notice that the size of \texttt{list\_data\_to\_gene\_checked} and the order of 1-D arrays in it should match the corresponding processes in the Generator kernel.

\begin{lstlisting}[language=Python]
def prediction_check(list_data_to_pred, list_data_to_gene):
"""
User defined predictions check function.
Check the predictions from Prediction processes (e.g. STD). 

Args:
    list_data_to_pred (list): list of data_to_pred gathered from all generators, sorted by the rank of generator.
                              Source: list of data_to_pred from UserGene.generate_new_data()
                              [1-D numpy.ndarray, 1-D numpy.ndarray, ...], size equal to number of generators.
    list_data_to_gene (list): list of data_to_gene gathered from all models in prediction kernel, sorted by the rank of model.
                              Source: data_to_gene_list from UserModel.predict()
                              [numpy.ndarray, numpy.ndarray, ...], array shape (n_pred, model output size), size equal to number of generators.

Returns:
    list_input_to_orcl (list): list of user defined input to oracle to generate ground truth.
                               Destination: list of input_for_orcl at UserOracle.run_calc().
                               [1-D numpy.ndarray, 1-D numpy.ndarray, ...]
    list_data_to_gene_checked (list): list of predictions distributed to generators.
                              Destination: list of data_to_gene to UserGene.generate_new_data(), length must match the number of generators and should be sorted by the rank of generator.
                              [1-D numpy.ndarray, 1-D numpy.ndarray, ...]
"""
list_input_to_orcl = []
list_data_to_gene_checked = []

##### User Part #####
threshold = 0.0  # set the threhold for standard deviation (std)

list_data_to_gene = np.array(list_data_to_gene, dtype=float)
std = np.std(list_data_to_gene, axis=1, ddof=1)  # calculate std of PL predictions
# identify PL input with high prediction std
i_orcl = np.where((std > threshold).any(axis=1))[0]
list_input_to_orcl = [list_data_to_pred[i] for i in i_orcl]

# limit the growth of list_input_to_orcl in this specific example to save memory
i = np.random.randint(0, 2)
list_input_to_orcl = list_input_to_orcl[:i]

pred_list = np.mean(list_data_to_gene, axis=1)  # take the mean of predictions to send to generator
pred_list[i_orcl] = 0  # for predictions with high std, send 0 instead to generator
list_data_to_gene_checked = list(pred_list)

return list_input_to_orcl, list_data_to_gene_checked
\end{lstlisting}

It might be desired to periodically adjust data waiting for labeling with the most up-to-date ML model to save the Oracle resources. This can be achieved by implementing the \texttt{adjust\_input\_for\_oracle} function that takes as inputs the list of data for labeling (\texttt{to\_orcl\_buffer}) and corresponding predictions from the most up-to-date ML models in the Training kernel (\texttt{pred\_list}). The order of inputs in the buffer could be altered and some inputs could be removed according to the predictions. This function is turned on by setting the \texttt{dynamic\_orcale\_list} to \texttt{True} in \texttt{al\_setting}.

\begin{lstlisting}[language=Python]
def adjust_input_for_oracle(to_orcl_buffer, pred_list):
"""
User defined function to adjust data in oracle buffer based on the corresponding predictions in pred_list.
Called only when dynamic_orcale_list is True in al_setting.

Args:
    to_orcl_buffer (list): list of input for oracle labeling.
                           Source: list of input_to_orcl to UserOracle.run_calc().
                           [1-D numpy.ndarray, 1-D numpy.ndarray, ...], size equal to number of elements in the oracle buffer
    pred_list (list): list of corresponding predictions of to_orcl_buffer from retrained ML.
                      Source: UserModel.predict()
                      [1-D numpy.ndarray, 1-D numpy.ndarray, ...], size equal to number of elements in the oracle buffer
Returns:
    to_orcl_buffer (list): list of adjusted input for oracle labeling. (list of input_to_orcl to UserOracle.run_calc())
                           Destination: list of input for oracle labeling.
                           [1-D numpy.ndarray, 1-D numpy.ndarray, ...]
"""

##### User Part #####
threshold = 0.0  # set the threhold for standard deviation (std)

std = np.std(np.array(pred_list, dtype=float), axis=0, ddof=1)  # calculation std of predictions from retrained ML
# sort the to_orcl_buffer list based on the std
i_orcl_sorted = np.argsort(np.mean(std, axis=1), axis=0)[::-1]
to_orcl_buffer = np.array(to_orcl_buffer, dtype=float)[i_orcl_sorted]
std = std[i_orcl_sorted]
to_orcl_buffer = list(to_orcl_buffer[np.nonzero((std > threshold).any(axis=1))[0]])  # remove data with prediction std not exceeding the threshold 

return to_orcl_buffer
\end{lstlisting}

\end{document}